\documentclass[10pt,twocolumn,letterpaper]{article}

\usepackage{wacv}
\usepackage{times}
\usepackage{epsfig}
\usepackage{graphicx}
\usepackage{amsmath}
\usepackage{amssymb}
\usepackage{xspace}
\usepackage{color}
\usepackage{multirow}
\usepackage{graphicx}
\usepackage{adjustbox}
\usepackage{subfigure}
\usepackage{paralist} 
\usepackage{enumitem} 
\usepackage{booktabs} 
\usepackage{array}
\usepackage{caption} 
\usepackage{mathrsfs}
\usepackage[pagebackref=true,breaklinks=true,colorlinks,bookmarks=false]{hyperref}

\usepackage{comment}

\graphicspath{{figures/}}

\newcommand{\Paragraph}[1]{\vspace{-2mm}{\flushleft{\textbf{#1}}}} 

\definecolor{gray}{rgb}{0.35,0.35,0.35}
\definecolor{MyBlue}{rgb}{0,0.2,0.8}
\definecolor{MyRed}{rgb}{0.8,0.2,0}
\definecolor{MyGreen}{rgb}{0.0,0.4,0.1}
\definecolor{MyGray}{rgb}{0.4,0.4,0.4}

\long\def\ignorethis#1{}

\newlength\paramargin
\newlength\figmargin
\newlength\secmargin

\newcolumntype{L}[1]{>{\raggedright\let\newline\\\arraybackslash\hspace{0pt}}m{#1}}
\newcolumntype{C}[1]{>{\centering\let\newline\\\arraybackslash\hspace{0pt}}m{#1}}
\newcolumntype{R}[1]{>{\raggedleft\let\newline\\\arraybackslash\hspace{0pt}}m{#1}}

\def\etal{et~al.\xspace}

\newcommand{\figref}[1]{Figure~\ref{fig:#1}}
\newcommand{\eqnref}[1]{\eqref{eq:#1}}

\newcommand{\method}[1]{\textsc{#1}}

\newcommand{\VQA}{VQA-1\xspace}

\newcommand{\VQATST}{{VQA 360$^\circ$}\xspace}
\newcommand{\cubavpool}{\method{Cubemap-Avgpool}\xspace}

\newcommand{\Tucker}{\method{Tucker}\xspace}
\newcommand{\TuckerM}{\method{Tucker\&Diffusion}\xspace}

\newcommand{\MLB}{\method{MLB}\xspace}

\newcommand{\centralcrop}{\method{Central-crop}\xspace}
\newcommand{\directcrop}{\method{Direct-split}\xspace}
\newcommand{\resize}{\method{Resize}\xspace}
\newcommand{\avpool}{\method{ResNet-Avgpool}\xspace}
\newcommand{\Qzero}{\method{Q-type prior}\xspace}

\wacvfinalcopy 

\def\wacvPaperID{****} 

\ifwacvfinal\fi
\begin{document}

\title{Visual Question Answering on 360$\mathbf{^\circ}$ Images}

\author{Shih-Han Chou$^{1,2}$, Wei-Lun Chao$^{3}$, Wei-Sheng Lai$^{5}$, Min Sun$^{2}$, Ming-Hsuan Yang$^{4,5}$\vspace{5pt}\\$^{1}$University of British Columbia \quad  $^{2}$National Tsing Hua University \quad  \\$^{3}$The Ohio State University \quad $^{4}$University of California at Merced \quad $^{5}$Google   
}

\twocolumn[{%
\renewcommand\twocolumn[1][]{#1}%
\maketitle
\begin{center}
    \captionsetup{type=figure}
	\footnotesize
	\vspace{-3em}
	\includegraphics[width=0.96\linewidth]{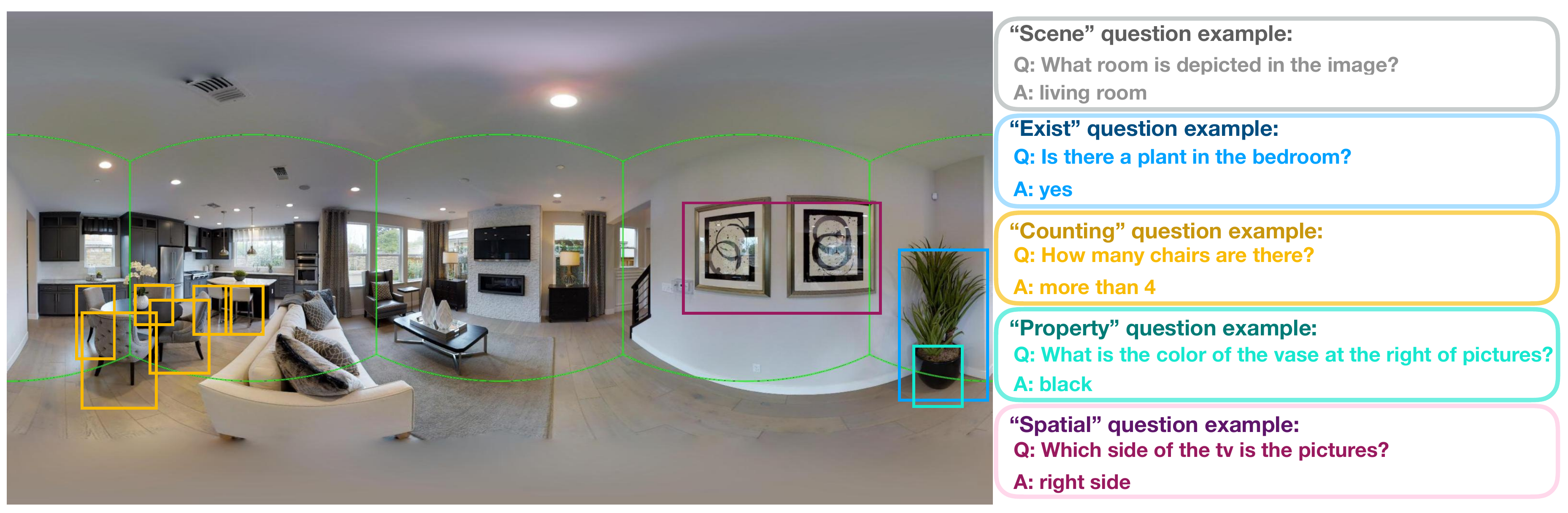}
	\vskip-8pt
	\captionof{figure}{
	    \textbf{An example of our \VQATST dataset.}
		We introduce \VQATST, a novel task of visual question answering on 360$^\circ$ images, and collect the first real \VQATST dataset, in which each image is annotated with around $11$ questions of five types (marked by different colors).
		The bounding boxes indicate where to look to infer the answers.
		%
		%
		Best viewed in color.
	}
	\label{fig:problem}
\end{center}
}]

\begin{abstract}
\vspace{-3mm}
In this work, we introduce \VQATST, a novel task of visual question answering on 360$^\circ$ images. 
Unlike a normal field-of-view image, a 360$^\circ$ image captures the entire visual content around the optical center of a camera, demanding more sophisticated spatial understanding and reasoning. 
To address this problem, we collect the first \VQATST dataset, containing around 17,000 real-world image-question-answer triplets for a variety of question types.
We then study two different VQA models on \VQATST, including one conventional model that takes an equirectangular image (with intrinsic distortion) as input and one dedicated model that first projects a 360$^\circ$ image onto cubemaps and subsequently aggregates the information from multiple spatial resolutions.
We demonstrate that the cubemap-based model with multi-level fusion and attention diffusion performs favorably against other variants and the equirectangular-based models.
Nevertheless, the gap between the humans' and machines' performance reveals the need for more advanced \VQATST algorithms.
We, therefore, expect our dataset and studies to serve as the benchmark for future development in this challenging task.
Dataset, code, and pre-trained models are available online.\footnote{http://aliensunmin.github.io/project/360-VQA/}
\vspace{-2mm}
\end{abstract}

\section{Introduction}
\label{sec:intro}

Visual question answering (VQA) has attracted significant attention recently across multiple research communities. 
In this task, a machine needs to visually perceive the environment, understand human languages, and perform multimodal reasoning---all of them are essential components to develop modern AI systems.
Merely in the past three years, more than two dozen datasets have been published, covering a wide variety of scenes, language styles, as well as reasoning difficulties~\cite{Antol_2015_ICCV, gao2015you, goyal2017making, johnson2017clevr, malinowski2014nips, ren2015exploring, zhu2016visual7w}. 
Together with those datasets are over a hundred algorithms being developed, consistently shrinking the gap between humans' and machines' performance~\cite{ben2017mutan, fukui2016multimodal, kafle2017analysis, kazemi2017show, kim2016hadamard}.

Despite such an explosive effort, existing work is constrained in the way a machine visually perceives the world. 
Specifically, nearly all the datasets use normal field-of-view (NFOV) images taken by consumer cameras.
Convolutional neural networks (CNNs) that are carefully designed for such images~\cite{he2016deep, VGG} have been necessary to extract powerful visual features. 
Nevertheless, NFOV images are not the only way, and very likely not the most efficient way, for a machine to interact with the world.
For example, considering a 360$^\circ$ horizontally surrounding scene, the NFOV of a consumer camera can only capture an $18\%$ portion~\cite{su2016activity}.
Such a fact, together with the reduced price of 360$^\circ$ cameras (e.g.,~Ricoh Theta~S, Samsung Gear~360, and GoPro~Omni), has motivated researchers to dig into 360$^\circ$ vision~\cite{cheng2018cube,chou2017self,hu2017deep,su2017learning}. 
We could imagine every robot to be equipped with a 360$^\circ$ camera in the near future.
It is thus desirable to extend VQA to such an informative visual domain.

In this work, we make the first attempt toward VQA on 360$^\circ$ images (\VQATST). 
Two major challenges immediately emerge. 
First, modern deep learning algorithms are heavily data consuming, yet so far, there is no publicly available dataset for \VQATST. 
Second, 360$^\circ$ (i.e., equirectangular) images have intrinsic distortion and larger spatial coverage, requiring a novel way to process visual inputs and perform sophisticated spatial reasoning.
Specifically, a machine needs to understand the spatial information in questions, search answers across the entire 360$^\circ$ scene, and finally aggregate the information to answer.

To resolve the first challenge, we collect the first real \VQATST dataset, using 360$^\circ$ images from real-world scenes.
Our dataset contains about 17,000 image-question-answer triplets with human-annotated answers (see an example in~\figref{problem}).
We have carefully taken the bias issue~\cite{goyal2017making, kafle2017analysis}, which many existing VQA datasets suffer, into account in designing our dataset.
We thus expect our dataset to benefit the development of this novel task.

\begin{figure}
\begin{center}
\includegraphics[width=0.95\linewidth]{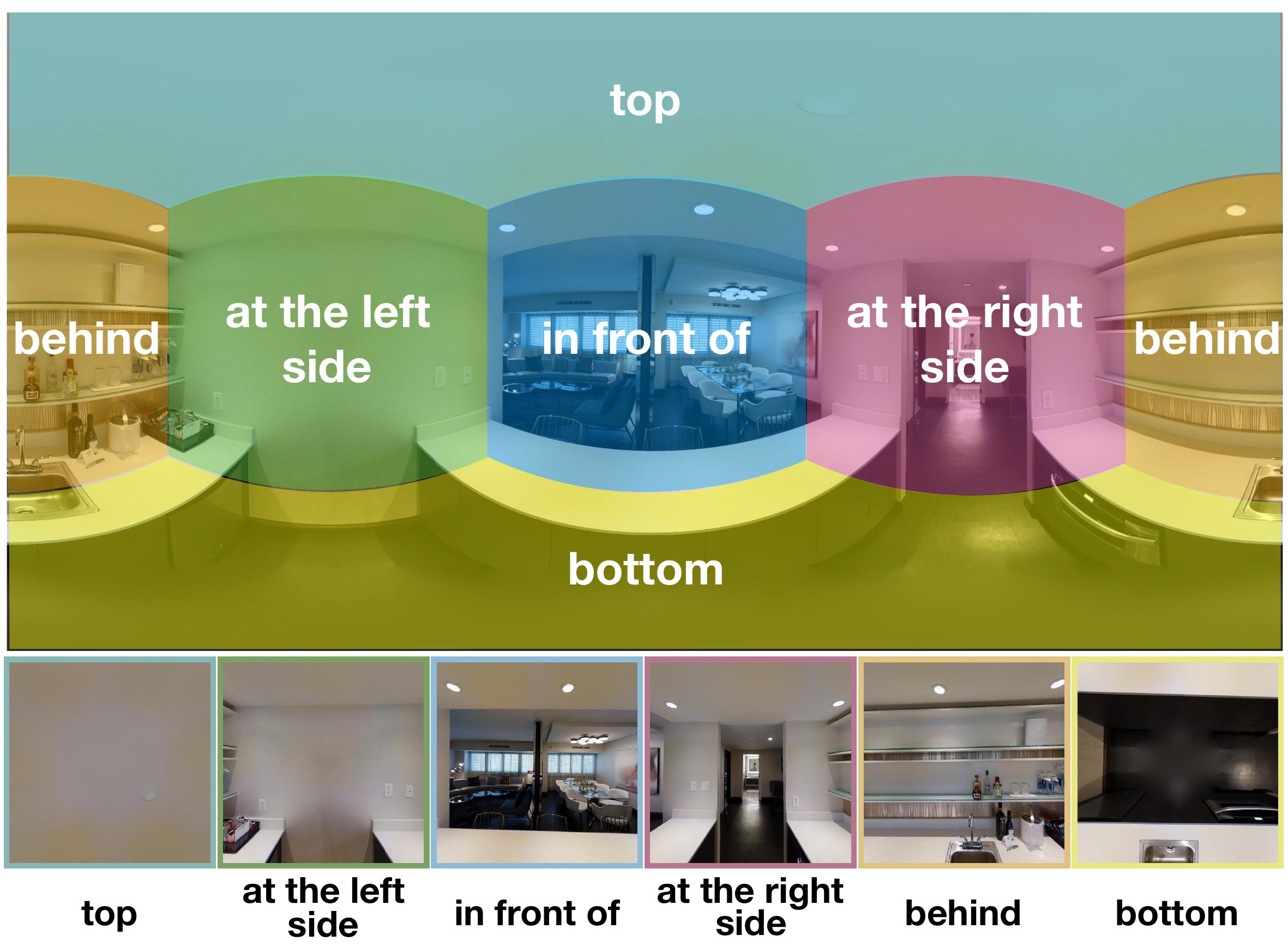}
\end{center}
\vskip -12.5pt
   \caption{\textbf{360$\mathbf{^\circ}$ image and cubemaps.}
        A equirectangular 360$^\circ$ image can be represented by six cubemaps, each corresponding to a spatial location, to reduce spatial distortion.}
\label{fig:tocube}
\vspace{-2mm}
\end{figure}

In addition, we study two models to address \VQATST. 
On the one hand, we use equirectangular images as input, similar to conventional VQA models on NFOV images.
On the other hand, to alleviate spatial distortion, we represent an input 360$^\circ$ image by six cubemaps~\cite{greene1986environment}.
Each map has its own spatial location and suffers less distortion (cf. Figure~\ref{fig:tocube}).
We develop a multi-level attention mechanism with spatial indexing to aggregate information from each cubemap while performing reasoning.
In this way, a machine can infer answers at multiple spatial resolutions and locations, effectively addressing the algorithmic challenge of \VQATST.
Moreover, cubemap-based architecture is flexible to take existing (pre-trained) VQA models as backbone feature extractors on cubemaps, effectively fusing multimodal information and overcoming the limited data issue.

We conduct extensive empirical studies to evaluate 
multiple variants of these models.
The superior performance by the cubemap-based model demonstrates the need to explicitly consider intrinsic properties of \VQATST, both visually and semantically.
By analyzing the gap between the machine's and the human's performance, we further suggest future directions to improve algorithms for \VQATST. 

Our contributions in this work are two-fold: 
\begin{compactitem}
\item We define a novel task named \VQATST. 
We point out the intrinsic difficulties compared to VQA on NFOV images.
We further collect the first real \VQATST dataset, which is designed to include complicated questions specifically for 360$^\circ$ images. 
\item We comprehensively evaluate two kinds of VQA models for \VQATST, including one that can effectively handle spatial distortion while performing multi-level spatial reasoning. We then point out future directions for algorithm design for \VQATST. 
\end{compactitem}

\section{Related Work}
\label{sec:related}

\vspace{1mm}
\Paragraph{VQA models.}
Visual Question Answering requires comprehending and reasoning with visual (image) and textual (question) information~\cite{zeng2017leveraging}.
The mainstream of model architectures is to first learn the joint image-question representation and then predict the answer through multi-way classification.
In the first stage, two mechanisms, \emph{visual attention}~\cite{anderson2018bottom, xu2015show,lu2016hierarchical} and \emph{multimodal fusion}~\cite{fukui2016multimodal,ben2017mutan}, have been widely explored.
For example, the stacked attention networks (SANs)~\cite{yang2016stacked} was developed to perform multi-round attention for higher-level visual understanding.
On the other hand, Fukui et al.~\cite{fukui2016multimodal} proposed the Multimodal Compact Bilinear pooling (MCB) to learn a joint representation, and Ben et al.~\cite{ben2017mutan} developed a tensor-based Tucker decomposition to efficiently parameterize the bilinear interaction.
Recently, several work~\cite{chen2019uniter,li2019visualbert,lu2019vilbert,su2019vl} extended BERT~\cite{devlin2018bert} by developing new pre-training tasks to learn (bidirectional) transformers~\cite{vaswani2017attention} for joint image and text representations.

Despite the variety of architectures, most of existing methods directly apply CNNs to the whole NFOV image to extract (local) features, which may not be suitable to 360$^\circ$ images.
In this paper, we explore a different architecture to extract CNN features from the cubemap representations of a 360$^\circ$ image and then fuse features across cubemaps.
The cubemap-based model shares some similarity to~\cite{anderson2018bottom, yang2016stacked}, yet we apply multiple-rounds of attentions to different spatial resolutions, one within and one across cubemaps, so as to achieve better spatial understanding.

\Paragraph{VQA datasets.}
There have been over two dozen of VQA datasets on NFOV images published in recent years.
Most of them aim for open-ended answering~\cite{Antol_2015_ICCV,goyal2017making,krishna2017visual}, providing for a pair of image and question with one or multiple correct answers~\cite{chao2018being, zhu2016visual7w}. 
An alternative setting is multiple-choice answering: a set of candidate answers are provided for each question, in which one of them is correct.
Our \VQATST dataset belongs to the first category but focuses on a very different input domain, 360$^\circ$ images.

We note that there are two emerging VQA tasks, embodied QA~\cite{das2018embodied} and interactive QA~\cite{gordon2018iqa}, that require a machine to interact with the 3D environment (e.g., turn right or move closer).
Our dataset and task are different, from two aspects.
First, we work on real-world scenes, while both of them are on synthetic ones.
Second, we take 360$^\circ$ images as input while they take NFOV images. A machine there has to take actions to explore the environment, being less efficient.

\Paragraph{360$\mathbf{^\circ}$ vision.}
With the growing popularity of virtual reality (VR) and augmented reality (AR), 
360$^\circ$ images and videos have attracted increasing attention lately.
One of the interesting problems is to automatically navigate a 360$^\circ$ video~\cite{hu2017deep,su2017learning,su2016activity} or create a fast-forward summary~\cite{lai2017semantic}.
Other research topics include 360$^\circ$ video stabilization~\cite{kopf2016360}, compression~\cite{su2018learning}, saliency prediction~\cite{cheng2018cube}, depth estimation~\cite{de2018eliminating}, and object detection~\cite{chou2019360, su2017learning}.
Recently, Chou~\etal~\cite{chou2017self} study visual grounding to localize objects in a 360$^\circ$ video for a given narrative, while Chen~\etal~\cite{chen2019touchdown} explore natural language navigation in 360$^\circ$ street environments.
In contrast to these tasks, VQA on 360$^\circ$ images requires further inferring the answers according to questions,  demanding more sophisticated reasoning of the scene.

\begin{table*}
\begin{center}
\footnotesize
\tabcolsep 2.6pt
\begin{tabular}{c|l|l|l}
\toprule
\multicolumn{1}{l|}{\textbf{Q type}} & \multicolumn{1}{c|}{\textbf{Template}} & \multicolumn{1}{c|}{\textbf{Example}} &\multicolumn{1}{c}{\textbf{Answer}} \\ 
\midrule
Scene & \begin{tabular}[c]{@{}l@{}}What room is depicted in the image? \end{tabular} & \begin{tabular}[c]{@{}l@{}}What room is depicted in the image?\end{tabular} & \begin{tabular}[c]{@{}l@{}} bedroom/... \end{tabular}\\ \midrule
Exist & \begin{tabular}[c]{@{}l@{}}Is/Are there (a) $<$obj1$>$ \textunderscore\textunderscore\textunderscore? \\ + in the $<$scene$>$ \\ + $<$direc$>$ \\ + $<$direc$>$ of the $<$obj2$>$ \\ + $<$direc$>$ of the $<$obj2$>$ in the $<$scene$>$ \end{tabular} & \begin{tabular}[c]{@{}l@{}} \\ Is there a chair in the kitchen? \\ Is there a chair at my right side? \\ Is there a chair at the right side of the window? \\ Is there a chair at the right side of the window in the kitchen? \end{tabular} & \begin{tabular}[c]{@{}l@{}} yes/no \end{tabular}\\ \midrule
Counting & \begin{tabular}[c]{@{}l@{}}How many $<$obj1$>$ are \textunderscore\textunderscore\textunderscore? \\ + in the $<$scene$>$ \\ + $<$direc$>$ \\ + $<$direc$>$ of the $<$obj2$>$ \\ + $<$direc$>$ of the $<$obj2$>$ in the $<$scene$>$ \end{tabular} & \begin{tabular}[c]{@{}l@{}} \\ How many chairs are in the kitchen? \\ How many chairs are at my right side? \\ How many chairs are at the right side of the window? \\ How many chairs are at the right side of the window in the kitchen? \end{tabular} & \begin{tabular}[c]{@{}l@{}} 0/1/2/... \end{tabular}\\ \midrule
Property & \begin{tabular}[c]{@{}l@{}}What is the ($<$color$>$) $<$obj1$>$ \textunderscore\textunderscore\textunderscore \hspace{.1em} made of? \\ What is the color of the $<$obj1$>$ \textunderscore\textunderscore\textunderscore?\\ + in the $<$scene$>$ \\ + $<$direc$>$ \\ + $<$direc$>$ of the $<$obj2$>$ \\ + $<$direc$>$ of the $<$obj2$>$ in the $<$scene$>$ \end{tabular} & \begin{tabular}[c]{@{}l@{}} \\ \\ What is the red sofa in the bedroom made of?\\ What is the red sofa at my right side made of? \\ What is the color of the sofa at the right of the window? \\ What is the color of the sofa at the right of the window in the bedroom? \end{tabular} & \begin{tabular}[c]{@{}l@{}} \\ \\ plastic/wood/... \\ red/brown/... \end{tabular}\\ \midrule
Spatial & \begin{tabular}[c]{@{}l@{}}Where can I find the \textunderscore\textunderscore\textunderscore \hspace{.1em}$<$obj1$>$? \\ Which side of the \textunderscore\textunderscore\textunderscore \hspace{.1em} $<$obj1$>$ is the \textunderscore\textunderscore\textunderscore  \hspace{.3em}$<$obj2$>$?\\ + $<$color$>$ \\ + $<$material$>$ \end{tabular} & \begin{tabular}[c]{@{}l@{}} \\ \\ Where can I find the white flowers?\\ Which side if the white chair is the wooden door? \end{tabular} & \begin{tabular}[c]{@{}l@{}} \\ \\ in front of you/... \\ right side/... \end{tabular}\\
\bottomrule
\end{tabular}
\end{center}
\vskip -10pt
\caption{\textbf{Question templates and examples.} 
        We design the following question templates and utilize the scene types and semantic segmentation of the images to automatically generate questions.}
\label{tab:Qtemplete}
\vskip -5pt
\end{table*}


\section{\VQATST Dataset}
\label{sec:dataset}

We first present the proposed \VQATST dataset to give a clear look at the task and its intrinsic challenges.
We begin with the dataset construction, including image collection, question generation, and answer annotation.
We then provide detailed statistics for our \VQATST dataset.

\subsection{Images Collection} \label{subsec:imagecollect}

We focus on indoor scenes as they are usually more dense with contents such as objects, which are suitable for developing algorithms for sophisticated reasoning.
In contrast, outdoor scenes, like those in~\cite{hu2017deep, lai2017semantic, su2018learning,su2016activity}, capture certain (ego-centric) activities and are of sparse contents, which are more suitable for summarization or navigation.

We collect 360$^\circ$ images of indoor scenes from two publicly accessible datasets, Stanford 2D-3D~\cite{2017arXiv170201105A} and Matterport3D~\cite{Matterport3D}. 
Both datasets provide useful side information such as scene types and semantic segmentation, which benefit question generation.
There are about $23$ different scenes, including common areas in houses (e.g., bathroom, kitchen, bedroom, etc.) and workplaces (e.g., office, conference room, auditorium, etc.).
To maximize the image diversity, we discard images captured in the same room but with different viewpoints.
In total, we collect $744$ images from the Stanford 2D-3D dataset and $746$ images from the Matterport3D dataset.

All the 360$^\circ$ images are stored in the equirectangular format and resized to $1024 \times 512$.
The equirectangular projection maps latitude and longitude of a sphere to the horizontal and vertical lines (e.g., a point at the top of the sphere is mapped to a straight line in an equirectangular image), which inevitably introduces heavy spatial distortion.

\subsection{Question Generation}
We design several question templates (c.f. Table~\ref{tab:Qtemplete}), together with the semantic segmentation and scene types associated with each 360$^\circ$ image\footnote{We can obtain room types and objects appearing in the scenes.}, to automatically generate questions.
Our templates contain five different types: ``scene'', ``exist'', ``counting'', ``property'' and ``spatial''. 
While imposing templates limit the diversity of questions, the main purpose of our dataset is to promote VQA on a new visual domain that has larger spatial coverage and complexity.
As illustrated in Figure~\ref{fig:problem}, a 360$^\circ$ image can easily contain multiple objects distributed at multiple locations.
\emph{We thus specifically design the question templates---either include spatial specifications or ask for spatial reasoning---to disambiguate the questions and encourage machines to acquire better spatial understanding.}
For instance, to answer ``What is the color of the vase at the right of pictures?'' in Figure~\ref{fig:problem}, a machine needs to first find the pictures (rightmost), look to the right to find the vase, and return the color\footnote{There are three vases in Figure~\ref{fig:problem}. Adding spatial specifications is thus necessary, and different specifications will lead to different answers.}.
To answer ``Which side of the TV is the pictures?'', a machine needs to detect the TV and picture, and then return their relative spatial information in the scene.
Both examples require visual and spatial understanding at multiple resolutions and locations, which are scarce in existing VQA datasets on NFOV images (see the supplementary material for details).
On average, we create $11$ questions per image.

\subsection{Answer Annotations \& Question Refinements}
\label{subsec:annotation}
We resort to human annotators to provide precise answers.
We ask $20$ in-house annotators to answer the questions in our dataset.
To avoid synonyms words and to ease the process, we offer candidate answers according to the question types for annotators to select directly.
Annotators can also type free-form answers if none of the candidates is applicable.
We note that the automatically generated questions might be irrelevant to the image or lead to ambiguous answers\footnote{For instance, if there are two chairs with different colors, a question ``What is the color of the chair?'' will lead to ambiguous answers.}.
In such cases, we instruct the annotators to slightly modify the questions---e.g., by adding spatial specifications---to make them image-related or identifiable.
We also instruct annotators to draw bounding boxes (for a subset of image-question pairs), which indicate specific objects or locations associated with the answer.
Such information facilitates the analysis of model performances.

\begin{table}
\begin{center}
\footnotesize
\begin{tabular}{c|ccc}
        \toprule
                    & Training & Validation & Test \\
        \midrule
        $\#$images  & 743   & 148  & 599 \\
        QA pairs    & 8227  & 1756 & 6962 \\
        $\#$unique answers & 51 & 51 & 53 \\
        \midrule
        $\#$Scene type Q & 765 & 150 & 614 \\
        $\#$Counting type Q  & 1986 & 495 & 1934 \\
        $\#$Existed type Q  & 2015 & 417 & 1655 \\
        $\#$Property type Q  & 1355 & 322 & 1246 \\  
        $\#$Spatial type Q  & 2106 & 372 & 1513 \\
        \bottomrule
    \end{tabular}
\end{center}
\vskip -12.5pt
\caption{\textbf{Summary of 360$\mathbf{^\circ}$ VQA dataset.}
        We summarize the number of images, QA pairs, and unique answers in each split of our dataset.
        We also provide a detailed statistic for each type of question.}
\label{tab:dataset_statistic}
\vskip -5pt
\end{table}

\begin{figure}
\begin{center}
\includegraphics[width=0.9\linewidth]{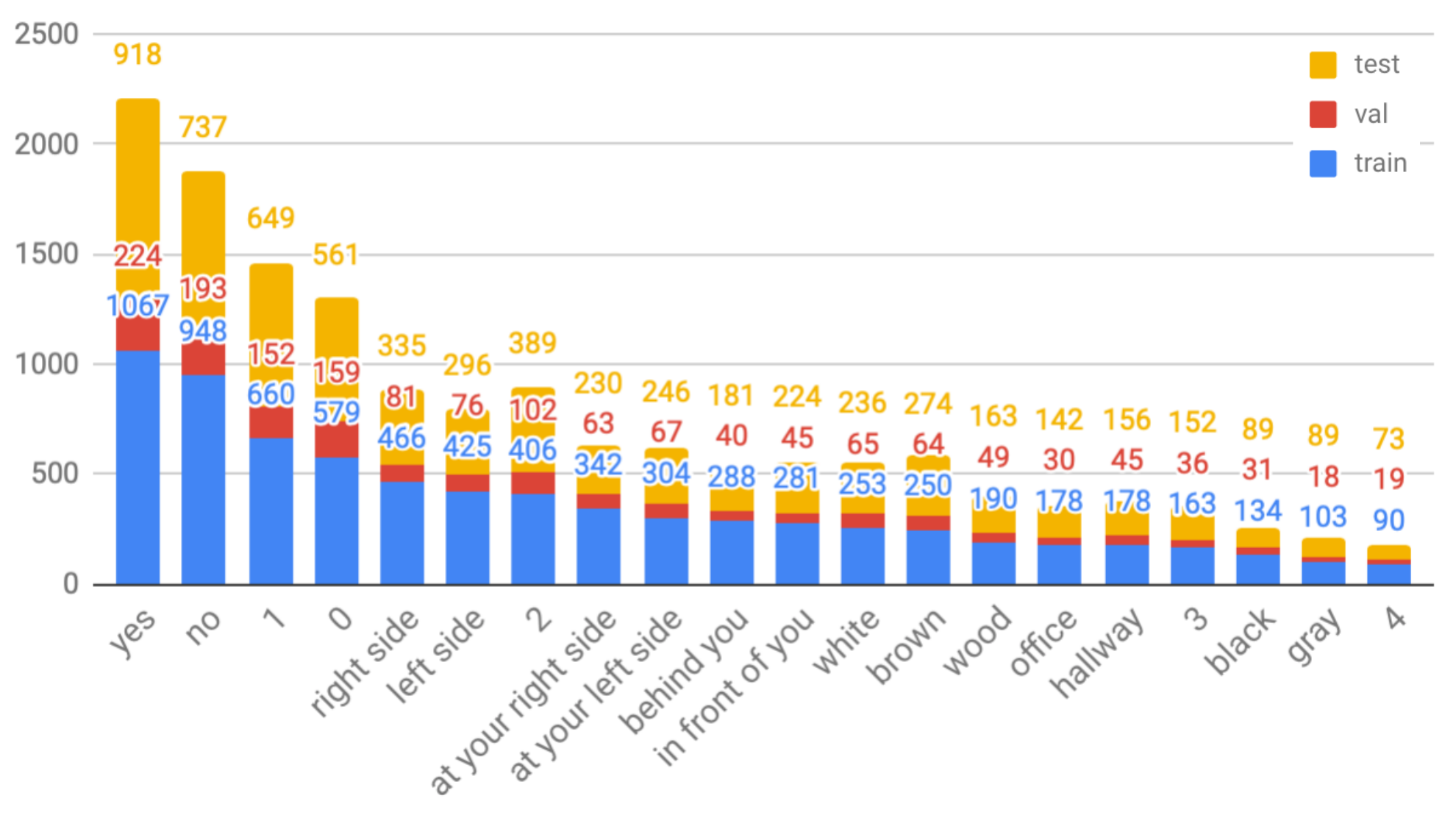}
\end{center}
\vskip -12.5pt
   \caption{
    \textbf{Distribution of answers.}
    We balance our dataset such that the answers of the same question type appear uniformly (e.g., ``yes/no'', ``0/1'', and ``right side/left side'').}
\label{fig:statistic}
\vskip -5pt
\end{figure}

\subsection{Dataset Statistics} 
\label{subsec:datasetstatistics}
 
Our \VQATST dataset consists of $1,490$ images and $16,945$ question-answer pairs, which are split into the training, validation, and test sets with $50\%$, $10\%$, and $40\%$ of images, respectively.
We summarize the statistics in Table~\ref{tab:dataset_statistic} and show the distribution of the top $20$ answers in~Figure~\ref{fig:statistic}. 
We note that each question type has at least $2$ corresponding answers in the top $20$ ones. 
Moreover, those from the same type have the similar number of presence (e.g., ``yes/no'', ``0/1'', ``right/left side''), preventing a machine from cheating by predicting the dominant answer.
For question types with a few unique answers, we make sure that the unique answers appear almost uniformly to minimize dataset bias.
%

\section{\VQATST Models}
\label{sec:algorithm}

\begin{figure*}
\begin{center}
\includegraphics[width=.85\linewidth]{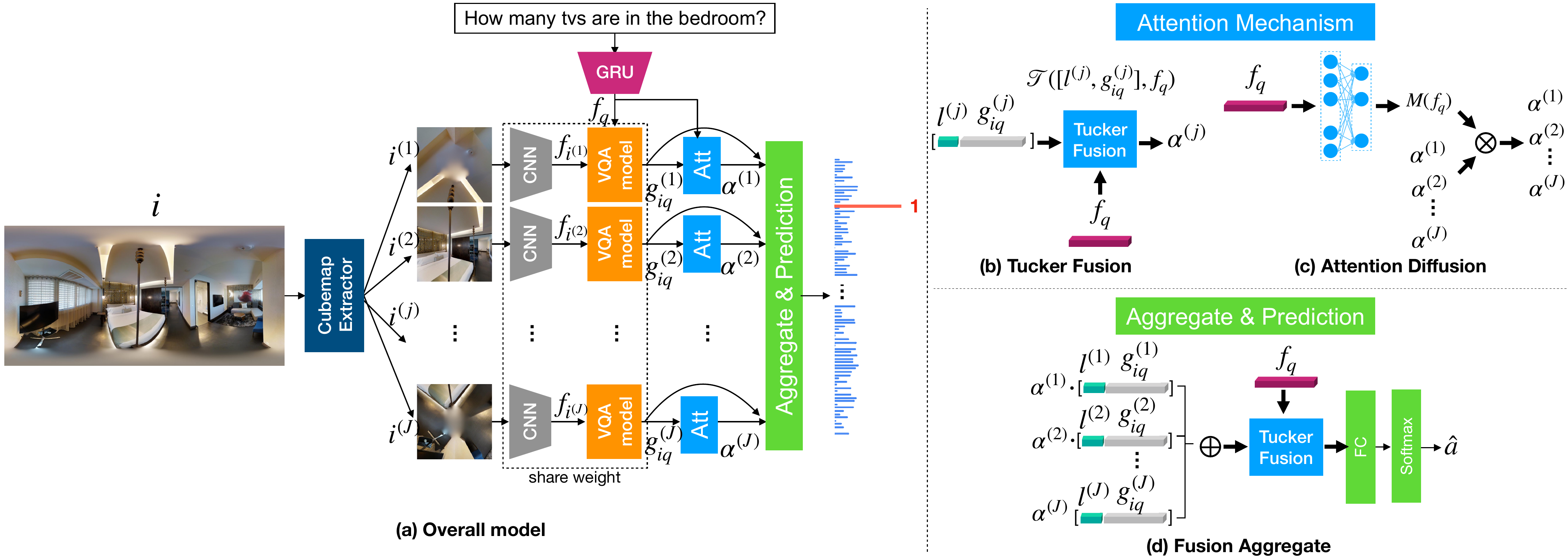}
\end{center}
\vskip-12.5pt
   \caption{\textbf{VQA 360$\mathbf{^\circ}$ models.}
    We propose a cubemap-based architecture that first extracts visual features from the cubemaps of the input 360$^\circ$ image and then performs bottom-up multi-level attention and feature aggregation.}
\label{fig:model}
\end{figure*}
In this section, we study two VQA models, including one dedicated to resolving inherent challenges in \VQATST.

\Paragraph{Notations and problem definitions.}
Given a question $q$ and an image $i$, a machine needs to generate the answer $a$.
One common VQA model is to first extract visual features $f_i = \mathcal{F}_I(i)$ and question features $f_q = \mathcal{F}_Q(q)$, followed by multimodal representations $g_{iq} = \mathcal{G}(f_i, f_q)$.
The multimodal representations are then inputted into a classifier $\mathcal{C}(\cdot)$ of $K$ classes, corresponding to the top $K$ frequent answers, to generate the answer $a$.
Representative choices for $\mathcal{F}_I(\cdot)$ and $\mathcal{F}_Q(\cdot)$ are CNN and RNN models~\cite{yang2016stacked}, respectively.

\subsection{Equirectangular-based Models}\label{subsec:equi}
As the most common format to store and display a 360$^\circ$ image is the equirectangular projection into a 2D array, we can indeed directly apply existing (pre-trained) VQA models for \VQATST.
We take the Multimodal Low-rank
Bilinear Attention Network (\MLB) model~\cite{kim2016hadamard}
as an example, which adopts an efficient bilinear interaction for $\mathcal{G}(f_i, f_q)$.
We first extract the visual features $f_i$ by the pre-trained ResNet-152~\cite{he2016deep} and adopt the Gated Recurrent Units (GRU)~\cite{chung2014empirical, kiros2015skip} to extract the question features $f_q$. 
We then input the resulting $g_{iq}=\mathcal{G}(f_i,f_q)$ into a fully-connected layer with $K$ output units to build a $K$-way classifier $\mathcal{C}(\cdot)$. 
We optimize the whole network using the training set of our \VQATST dataset and set $K$ to be the number of unique training answers (i.e., $51$).

The \MLB model $\mathcal{G}(f_i,f_q)$ pre-trained on the \VQA~\cite{Antol_2015_ICCV} dataset requires $f_i$ to retain a $14\times 14$ spatial resolution, equivalent to inputting a $448\times 448$ image to the ResNet.
We thus adopt a few strategies, including cropping or resizing the original 360$^\circ$ image, or inputting the original image while resizing the output ResNet features into a $14\times 14$ spatial resolution by an average pooling layer.
We analyze these strategies in Section~\ref{sec:exp}.

\Paragraph{Challenges.}
While the above strategies allow us to exploit VQA models pre-trained on much larger NFOV datasets (e.g., \VQA~\cite{Antol_2015_ICCV}), applying CNNs directly on 360$^\circ$ images suffers the inherent spatial distortion~\cite{su2017learning}.
On the other hand, adopting specifically designed spherical convolutions~\cite{su2017learning} prevents us from leveraging existing models and pre-trained weights.
An intermediate solution that takes both concerns into account is thus desirable.

Moreover, existing VQA models like \MLB~\cite{kim2016hadamard} and SAN~\cite{yang2016stacked} only consider a single visual resolution when performing feature aggregation in $\mathcal{G}(f_i, f_q)$.
For 360$^\circ$ images that cover a large spatial range, a more sophisticated mechanism that involves multiple resolutions of feature aggregation is required.
To this end, we propose a cubemap-based model to simultaneously tackle the above challenges.

\subsection{Cubemap-based Models}
To reduce spatial distortion, we first represent a 360$^\circ$ image by six non-overlapping cubemaps, $\{i^{(j)}\}_{j=1}^J$, via the perspective projection (c.f. Figure~\ref{fig:tocube}; see the supplementary material for details).
Each cubemap corresponds to a specific portion of the 360$^\circ$ image with less distortion.
Collectively, the cubemaps together can recover the original image.
This representation naturally leads to a bottom-up architecture that begins with the local region understanding and then global reasoning (cf. Figure~\ref{fig:model}).

In the first stage, we can apply any existing VQA models, e.g., \MLB~\cite{kim2016hadamard}, to each cubemap individually, resulting in $J$ local multimodal representations:
\begin{align}
g_{iq}^{(j)}=\mathcal{G}(f_{i^{(j)}}, f_q)\,,
\end{align}
where $f_{i^{(j)}}$ denotes the visual features of the $j$-th cubemap. 

\Paragraph{Bottom-up multi-level attention.}
In the second stage, the main challenge is to effectively aggregate information from cubemaps.
While average and max pooling have been widely used, they simply ignore the location associated with each cubemap.
We thus resort to the attention mechanism:
\begin{align}
 & g_{i} = \sum_{j=1}^J \alpha^{(j)}g_{iq}^{(j)},\hspace{10pt}\text{s.t.}~\alpha^{(j)}\geq 0 ,~ \sum_j \alpha^{(j)} = 1. \label{eq:att}
\end{align}
The attention weight $\alpha^{(j)}$ can be computed according to information of each cubemap, \emph{including its location}, making aggregation more flexible.
As many existing VQA models already apply the attention mechanism \textit{within} the input images~\cite{kim2016hadamard,yang2016stacked} (e.g., a cubemap in our cases), the attention to aggregate \textit{across} cubemaps is actually the second-level of attention but on a coarse resolution. 

We apply Tucker fusion $\mathcal{T}(\cdot,\cdot)$~\cite{ben2017mutan}  to compute the attention weights according to the cubemap feature $g_{iq}^{(j)}$, location indicator $l^{(j)}$, and question feature $f_q$: Tucker fusion has been shown effective and efficient in fusing information from multiple modalities. The resulting $\alpha^{(j)}$ is as follows,
\begin{equation}
    \alpha^{(j)} =\text{softmax}\{\mathcal{T}([l^{(j)}, g_{iq}^{(j)}], f_q)\}\,, \label{eq:Tucker}
\end{equation}
where $[\cdot,\cdot]$ means concatenation. The softmax is performed over $j\in\{1,\cdots,J\}$. We use a one-hot vector $l^{(j)}$ to encode the cubemap location. 
In this way, the attention weights can zoom into the cubemap location mentioned in the question.

\Paragraph{Attention diffusion.} 
The attention weighs by~\eqnref{Tucker}; however, do not explicitly consider spatial relationship across cubemaps.
For a question like ``Is there a chair at the right side of the window?'', we would expect the model to first attend to the cubemap that contain the window, and then \emph{shift} its attention to the cubemap
at the right.
To incorporate such a capability, we learn a diffusion matrix $M(f_q)$ conditioned on the question $f_q$: the entry $M(f_q)_{u,v}$ indicates how much attention to be shifted from the cubemap $v$ to $u$.
The resulting formula for $g_i$ in~\eqnref{att} becomes:
\begin{align}
    g_{i}=\sum_{u=1}^J \left(\sum_{v=1}^J M(f_q)_{u,v}\alpha^{(v)}\right) g_{iq}^{(u)}, 
    \text{s.t.} \sum_{u=1}^J M(f_q)_{u,v}=1.
    \label{eq:TuckerM}
\end{align}
\vspace{-10pt}
\Paragraph{Answer prediction.}
The resulting feature $g_i$ in~\eqnref{TuckerM} or \eqnref{att} then undergoes another Tucker fusion to extract higher-level image-question interactions before inputted into the classifier $\mathcal{C}(\cdot)$.
We can also replace $g_{iq}^{(j)}$ in~\eqnref{TuckerM} or~\eqnref{att} by the concatenation of $g_{iq}^{(j)}$ and $l^{(j)}$ to incorporate location cues into $g_i$.
This strategy is, however, meaningless to average or max pooling---it simply results in an all-one vector. 
We illustrate the overall model architecture in~\figref{model}. More details are included in the supplementary material.

\section{Experimental Results}
\label{sec:exp}

\subsection{Setup}
\vspace{1mm}
\Paragraph{Variants of cubemap-based models.}
The cubemap-based model can take any existing VQA model as the backbone.
We choose the \MLB model~\cite{kim2016hadamard}, a bilinear multimodal fusion and attention model.
We experiment with other VQA backbones~\cite{ben2017mutan,singh2018pythia} in the supplementary material to demonstrate the applicability of the cubemap-based models.

We remove the fully-connected layer of the original \MLB model to extract multimodal features.
We apply the pre-trained \MLB model to each cubemap of size $448\times 448$, and consider the following three different \emph{aggregation schemes} before performing the final answer prediction.
\begin{compactitem}
    \item \cubavpool: apply average pooling on $g_{iq}^{(j)}$.
    \item \Tucker: attention weights by Tucker fusion in~\eqnref{Tucker}.
    \item \TuckerM: attention weights by Tucker fusion followed by the diffusion in~\eqnref{TuckerM}.
\end{compactitem}

\Paragraph{Variants of equirectangular-based models.}
We consider four ways to apply \MLB on the equirectangular images.
\begin{compactitem}
    \item \centralcrop: resize the shorter size of the image to $448$ to preserve the aspect ratio and then crop the image to $448 \times 448$ to extract ResNet features.
    \item \resize: resize the image into $448 \times 448$ without any cropping and extract ResNet features.
    \item \avpool: resize the shorter size of the image to $448$ and apply an average pooling layer on the ResNet output to obtain $14 \times 14$ resolution features.
    \item \directcrop: split an equirectangular image into $2 \times 3$ patches, resize each to $448 \times 448$ and apply \MLB, and then apply \TuckerM to aggregate information for predicting the answer.
\end{compactitem}
Note that the \directcrop and \TuckerM models have the same architecture but different inputs.

\Paragraph{Baselines.} 
We provide \Qzero, a model that outputs the most frequent answer of each question type.
\begin{table*}
\begin{center}
\footnotesize
\begin{tabular}{c|c|cc|ccccc}
        \toprule
        Model       & Variants & Overall avg & Avg by type & Scene & Exist & Counting & Property & Spatial\\ 
        \midrule
        \Qzero       &       -    & 33.50 & 31.71 & 25.41 & 55.47 & 33.56 & 21.99 & 22.14\\
        \midrule
        Equirectangular-based   & \centralcrop & 53.39 & 54.07 & 60.66 & 75.00 & 47.10 & 50.16 & 37.45\\
        Equirectangular-based   & \resize       & 54.21 & 55.77 & 68.46 & 75.66 & \textbf{47.31} & 51.48 & 35.96\\
        Equirectangular-based   & \avpool      & 54.47 & 56.14 & 69.34 & 76.81 & 46.32 & 50.96 & 37.25\\
        Equirectangular-based$^\star$  & \avpool & 54.15 & 55.55 & 67.48 & 77.17 & 46.17 & 49.04 & 37.90\\
        \midrule
        Equirectangular-based & \directcrop & 54.77 & 56.59 & 71.36 & 75.75 & 46.68 & 49.56 & 39.62\\
        \midrule
        Cubemap-based & \cubavpool & 54.60 & 56.23 & 69.17 & 76.22 & 46.79 & \textbf{51.72} & 37.26\\
        Cubemap-based & \Tucker & 57.71 & 59.07 & 69.89 & \textbf{77.23} & 46.53 & 48.24 & 53.47\\
        Cubemap-based & \TuckerM & \textbf{58.66} & \textbf{60.26} & \textbf{72.01} & 76.34 & 46.84 & 50.12 & \textbf{55.98}\\
        Cubemap-based$^\star$ & \TuckerM & 54.09 & 55.54 & 67.65 & 76.16 & 45.91 & 48.60 & 39.39 \\
        \bottomrule
    \end{tabular}
\end{center}
\vskip-10pt
\caption{\textbf{Quantitative results on the \VQATST test set.}
        The $^\star$ models are trained from scratch on the \VQATST training set without pre-training on the \VQA.
        The best result of each column is marked by the bold black color.}
\label{tab:exp}
\vspace{-3mm}
\end{table*}
\Paragraph{Implementation details.}
We first pre-train the backbone \MLB model on the \VQA~\cite{Antol_2015_ICCV} dataset, which contains over $100,000$ NFOV images and $300,000$ question-answer pairs for training.
Then, we plug the pre-trained model in all the compared models and fine-tune the models on our \VQATST training set for $150$ epochs. 
We optimize our models with the ADAM~\cite{kingma2014adam} optimizer and select the model with the best performance on the validation set.
\Paragraph{Evaluation metric.}
We use the top-1 accuracy for evaluation.
We report two types of accuracy: the average accuracy 
i) over all the questions, and
ii) over question types.

\subsection{Analysis and Discussions}
Table~\ref{tab:exp} summarizes the results on \VQATST test set.
The cubemap-based model with \TuckerM for attention weights performs favorably against other models, demonstrating the effectiveness of multi-level and \emph{diffused} attention on top of cubemaps representation for \VQATST.
In the following, we discuss several key observations.
\Paragraph{Limited language bias.}
The top row (Q-type prior) in Table~\ref{tab:exp} examines the dataset bias, which predicts the most frequent answer of each question type.
The inferior results suggest a low language bias in our dataset.
Specifically, for ``exist'' type questions that only have two valid answers each (i.e, ``yes'' or ``no''), using language prior is close to random guess.
Machines need to rely on images to answer.

\Paragraph{Equirectangular-based models.}
As shown in Table~\ref{tab:exp}, the \avpool model outperforms the \centralcrop and \resize, indicating the poor applicability of cropping and resizing to 360$^\circ$ images.
Since 360$^\circ$ images have large spatial coverage, in which objects might be of small sizes, resizing will miss those small objects while central cropping will lose $50\%$ of the image content.

\Paragraph{Cubemaps v.s. Equirectangular input.}
One major issue of applying existing VQA models directly to the 360$^\circ$ images is the spatial distortion. 
This is justified by the fact that all the equirectangular-based models are outperformed by all the cubemap-based models (except the \cubavpool one) on the overall performance.
Specifically, by comparing the \directcrop and \TuckerM, whose main difference is the input, the $3\sim 4\%$ performance gap clearly reflects the influence of distortion.
By looking into different question types, we also observe consistent improvements by applying cubemaps.

\Paragraph{Pre-training.}
Comparing the models with $\star$ (trained from scratch) and without $\star$ (with pre-training), the pre-trained weights (from the VQA-1 dataset) benefits the overall performance, especially for the cubemap-based models.

\Paragraph{Attention.}
Applying cubemaps resolves one challenge of \VQATST: spatial distortion.
We argue that a sophisticated way to aggregate cubemaps features to support spatial reasoning is essential to further boost the performance.
This is shown from the improvement by \TuckerM or \Tucker, compared to \cubavpool: the former two apply attention mechanisms guided by questions and cubemap locations for multi-level attention.
Specifically, \TuckerM outperforms \cubavpool by a notable $3.4\%$ at Avg. by Q type, mostly from the ``spatial'' question type. \TuckerM with spatial \emph{diffusion} also outperforms \Tucker in all the question types.
\begin{figure*}
\centering
\includegraphics[width=0.83\linewidth]{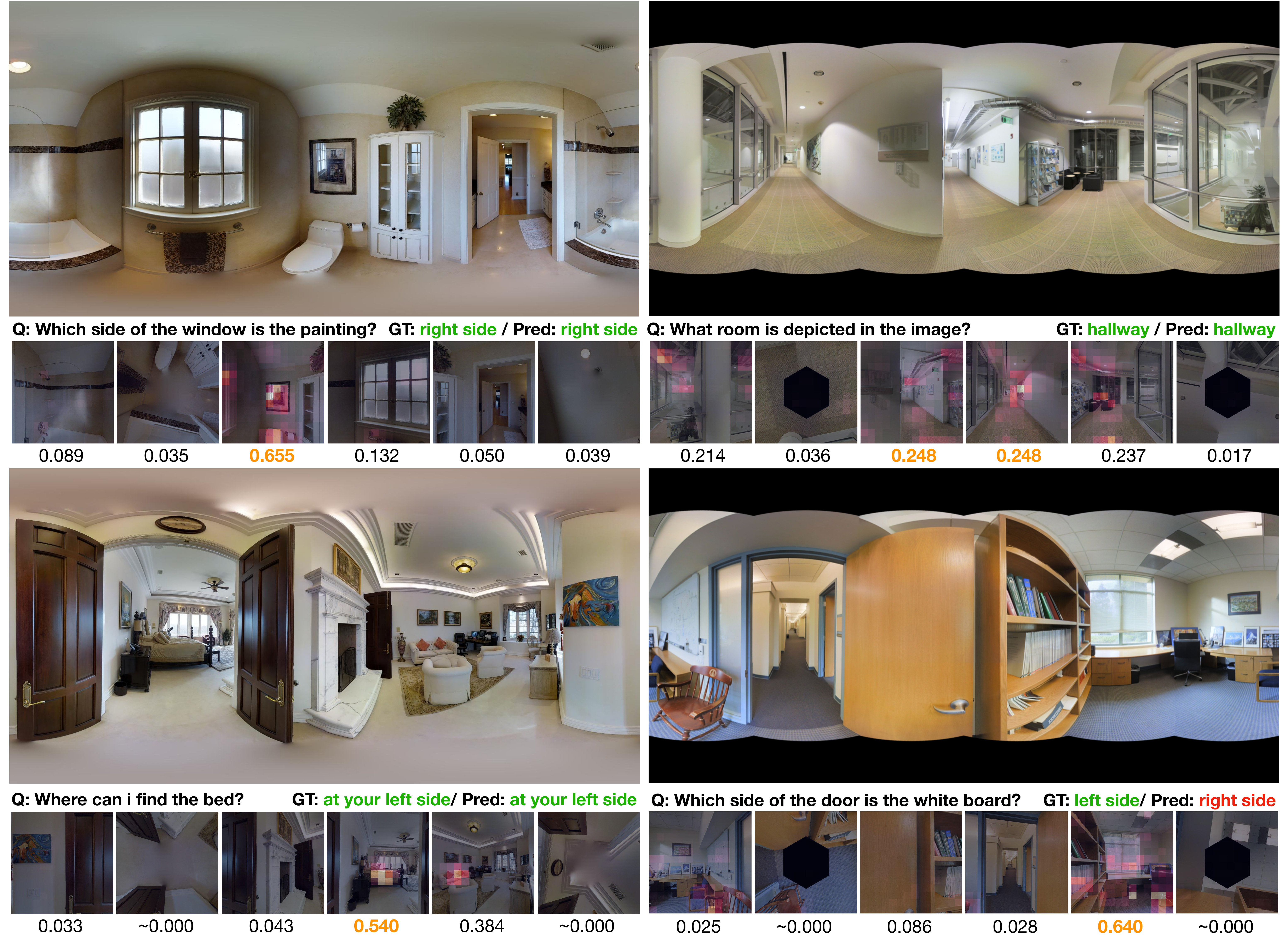}
\vspace{-2mm}
   \caption{\textbf{Visualization of attention.}
         We use the cubemap-based model \TuckerM as it performs the best.
         The digits below the cubemaps indicate the attention across cubemaps.
         The heat maps indicate the attention within cubemaps.}
\label{fig:result2}
\vspace{-5mm}
\end{figure*}
\begin{table}
\begin{center}
\footnotesize
\begin{tabular}{c|ccc}
        \toprule
        Model         & Avg. & Avg. by Q type & Spatial \\
        \midrule
        \Tucker (w/o) & 53.81 & 53.81 & 36.09 \\
        \Tucker (w/) & 57.71 & 59.07 & 53.47 \\
        \TuckerM  (w/o) & 54.91 & 56.51 & 39.13\\
        \TuckerM (w/) & 58.66 & 60.26 & 55.98\\
        \bottomrule
    \end{tabular}
\end{center}
\vskip-12.5pt
\caption{\textbf{Comparison of w/ and w/o location feature.}}
\label{tab:exp_indicator}
\vspace{-3mm}
\end{table}

\begin{table}
\begin{center}
\footnotesize
\renewcommand{\tabcolsep}{5pt} 
\begin{tabular}{c|c|ccccc}
        \toprule
         Model        & Overall & Scene & Exist & Counting & Property & Spatial\\
        \midrule
        Human & 84.05 & 88.95 & 91.79 & 71.58 & 89.97 & 85.25\\
        Machine & 59.80 & 68.89 & 77.12 & 49.65 & 45.81 & 61.97\\
        \bottomrule
    \end{tabular}
\end{center}
\vskip-10pt
\caption{\textbf{Results of human evaluation.} We also include the machine's performance on the same 1,000 questions to analyze the humans' and machines' gap.}
\label{tab:exp_human}
\vspace{-2mm}
\end{table}

\Paragraph{Location feature.}
Concatenating $l^{(j)}$ with $g_{iq}^{(j)}$ in~(\ref{eq:att}) and~(\ref{eq:TuckerM}) enables our model to differentiate cubemaps.
Table~\ref{tab:exp_indicator} compares the \TuckerM and \Tucker with/without $l^{(j)}$. 
The location indicator leads to consistent improvement, especially on the ``spatial'' type questions.

\Paragraph{Human Evaluation.}
We conduct a user study on our \VQATST dataset.
We sample $1,000$ image-question-answer triplets from the test set and ask at least two different users to answer each question. 
To ease the process, we give users five candidate answers, including the correct answer and four other answers that are semantically related to the question.
There are a total of $50$ unique users participating in the user study.
We note that the annotators labeling our dataset are not involved in the human evaluation to avoid any bias.

We summarize the results of human evaluation and the machine's prediction\footnote{We use our best cubemap-based model \TuckerM.}
in Table~\ref{tab:exp_human}.
Humans achieve a $84.05\%$ overall accuracy, which is at the same level as many existing VQA datasets~\cite{Antol_2015_ICCV, chao2018being, yu2015visual} and is much higher than another dataset on indoor images~\cite{malinowski2014nips}, justifying the quality of our \VQATST dataset.
Among the five question types, humans perform relatively poorly on ``counting'', which makes sense due to the complicated contents of $360^\circ$ images and the possible small objects.
Overall, there is about $\sim 25\%$ performance gap between human and machines. 
The gap is larger especially on ``counting'', ``property'', and ``spatial'' types, suggesting the directions to improve algorithms so as to match humans' inference abilities.

\Paragraph{Qualitative results.}
We present qualitative results in Figure~\ref{fig:result2}. 
Besides showing the predicted answers, we visualize the attention weights across cubemaps (by the digits) and within cubemaps (by the heat maps). 
The cubemap-based model with \TuckerM can zoom in to the cubemaps related to the questions, capture the answer regions, and aggregate them to predict the final answers.
Take the question ``Which side of the window is the painting?'' for example (the top-left one of~\figref{result2}). 
The model puts high attention on the cubemaps with windows and pictures and is able to infer the relative location.
For the question ``What room is depicted in the image?'' (the top-right of~\figref{result2}), the model distributes attention to all cubemaps except the top and bottom ones to learn information through them.
We also show a failure case in the bottom-right of~\figref{result2}. The question asks ``Which side of the door is the whiteboard?''.
However, the model mistakenly recognizes the window as the white board and incorrectly answers ``right side''.

\vspace{-2mm}
\section{Discussion and Conclusion}
\label{sec:conclusion}
\vspace{-2mm}

We introduce \VQATST, a novel VQA task on a challenging visual domain, 360$^\circ$ images. 
We collect the first \VQATST dataset and experiment with multiple VQA models. 
We then present a multi-level attention model to effectively handle spatial distortion (via cubemaps) and perform sophisticated reasoning.
Experimental results demonstrate the need to explicitly model intrinsic properties of 360$^\circ$ images, while the noticeable gap between humans' and machines' performance reveals the difficulty of reasoning on 360$^\circ$ images compared to NFOV images. 

We surmise that the gap may partially be attributed to the hand-crafted cubemap cropping. On one end, objects appear around the cubemap boundaries may be splitted. On the other end, it requires specifically designed mechanisms (e.g., attention diffusion~(\ref{eq:TuckerM})) to reason the spatial relationship among cubemaps. These issues likely explain the human-machine gap at the ``counting'' and ``spatial'' questions. Thus, to advance \VQATST, we suggest developing image-dependent cropping that detects objectness regions from the equirectangular images.
We also suggest developing a back-projection-and-inference mechanism that back-projects the detected objects into the 360$^\circ$ environment and performs reasoning accordingly.
Besides, the current questions are generated (or initialized) by templates. A future work is to include more human efforts to increase the question diversity. 
We expect our dataset and studies to serve as the benchmark for the future developments.

\Paragraph{Acknowledgments.}
This work is supported in part by NSF CAREER (\#~1149783) and MOST 108-2634-F-007-006 Joint Research Center for AI Technology and All Vista Healthcare, Taiwan.


\section{Supplementary Material}

In this section, we present additional results to complement the main paper.
\begin{compactitem}
    \item Section~\ref{sec:data_suppl}: Details on data collection (cf. Section 3 in the main paper).
    \item Section~\ref{sec:approach_suppl}: Implementation details of the proposed model (cf. Section 4.2 and 5.1 in the main paper).
    \item Section~\ref{sec:addition_exp}: Additional experimental results on the backbone VQA model and the answer prediction strategy for the cubemap-based models (cf. Section 5 in the main paper).
    \item Section~\ref{sec:result_suppl}: Additional qualitative results (cf. Section 5.3 in the main paper).
\end{compactitem}

\subsection{Data Collection} \label{sec:data_suppl}

\paragraph{Question generation.}
We design templates with place holders (cf. $<\cdot>$ in Table 2 of the main paper) to automatically generate questions.
We fill in $<$obj$>$ and $<$scene$>$ according to the semantic segmentation and scene types given by the Stanford 2D-3D~\cite{2017arXiv170201105A} and Matterport3D~\cite{Matterport3D} datasets. 
We fill in $<$color$>$ of $<$obj$>$ according to the corresponding pixel values.
For $<$direc$>$ of $<$obj$>$, we derive it from the corresponding cubemap location.
To generate questions with either ``no'' or ``0'' as the answer, we fill in the combinations of $<$obj$>$, $<$color$>$, and $<$direc$>$ that are not shown in the images.

\Paragraph{Answer annotation.}
We provide specific guidance (cf. Figure 2 of the main paper) for the annotators to identify the direction and location in a $360^\circ$ image.
We note that human annotators are allowed to modify the questions to make them less ambiguous or more related to the image contents.
In details, we instruct human annotators to modify the questions by following the templates in Table 1 in the main paper.
This flexibility also increases the diversity of the questions in our \VQATST dataset.

\paragraph{Question types.}
Our templates can be categorized into five different types: ``scene'', ``exist'', ``counting'', ``property'' and ``spatial''. 
\begin{compactitem}
\item ``Scene'' type: related to scene or room types, e.g., kitchen, office, etc.
\item ``Exist'' type: related to object presences and positions. 
\item ``Counting' type: for object counting and may involve object attributes and positions. 
\item ``Property'' type: for object attributes, e.g., color and material.
\item ``Spatial'' type: related to objects' relative positions and the photographer's relative position. 
\end{compactitem}
The ``exist'', ``counting'', ``property'', and ``spatial'' type questions generally require a model to infer answers from multiple locations (potentially across the entire scene) in a $360^\circ$ image.

\begin{figure*}[ht]
\centering
  \includegraphics[width=.7\linewidth]{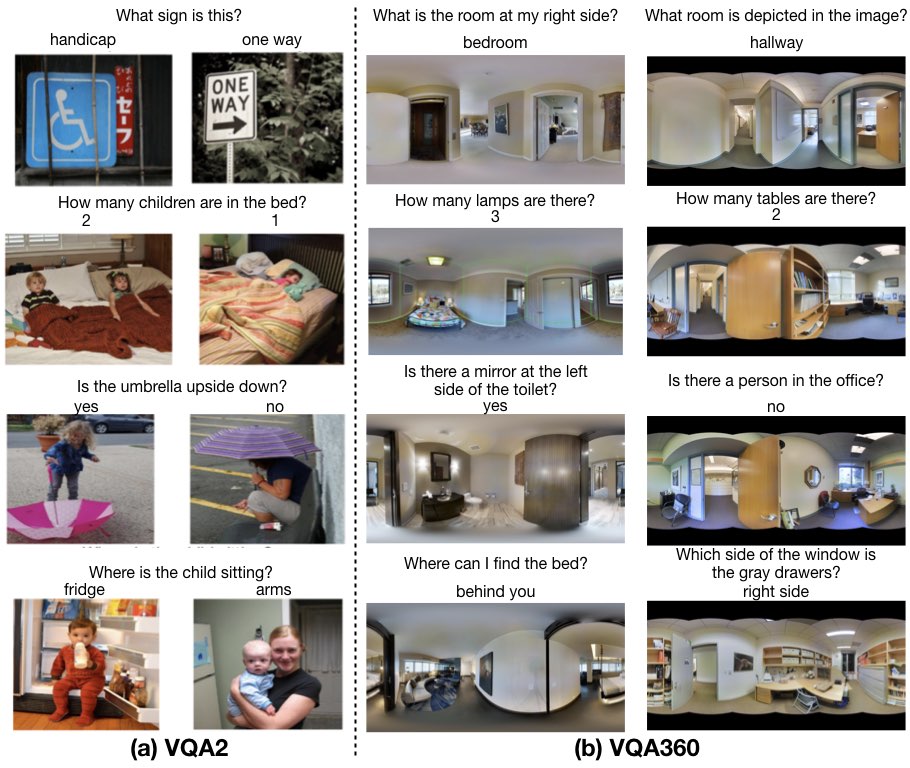}
  \caption{Examples of VQA2 and VQA 360$^\circ$.}
  \label{fig:rebuttal}
\end{figure*}

\paragraph{NFOV vs. 360$^\circ$ images.}
The demanding of spatial reasoning plays the key difference for Visual QA on 360$^\circ$ images (as mentioned in Section 3.2 in the main paper).
Therefore, the proposed dataset includes questions for spatial reasoning or with spatial cues, either by the templates or by the annotators.
Figure~\ref{fig:rebuttal} shows examples used in the VQA2 paper~\cite{goyal2017making}.
This is also evidenced by less than $3\%$ of the questions belonging to the ``where'' type in VQA2.
In contrast, objects in 360$^\circ$ images are highly distributed (even behind the observer), and we have more than $15\%$ ``where'' type questions.

\subsection{Implementation Details}
\label{sec:approach_suppl}

We provide the implementation details of the proposed model, following the notations introduced in Section 4.1 in the main paper.
We focus on \TuckerM, together with \MLB~\cite{kim2016hadamard} as the backbone VQA model and fusion aggregation to predict the answer (cf. Section 4.2 and Figure 4 in the main paper).
%
%
\begin{compactitem}
    \item {Step 1}: Extract the question feature $f_q = \mathcal{F}_Q(q)$.
    \item {Step 2}: Extract the visual feature $f_{i^{(j)}} = \mathcal{F}_I(i^{(j)})$ for each cubemap $j$.
    \item {Step 3}: Extract the local multimodal feature $g_{iq}^{(j)}=\mathcal{G}(f_{i^{(j)}}, f_q)$ for each cubemap $j$, where $\mathcal{G}(\cdot,\cdot)$ is the \MLB model without the last fully-connected layer.
    \item {Step 4}: Compute the attention weight $\alpha^{(j)}$ for each cubemap $j$, $\alpha^{(j)} \propto \exp\{\mathcal{T}([l^{(j)}, g_{iq}^{(j)}], f_q)\}$, where $\mathcal{T}$ is the Tucker fusion module~\cite{ben2017mutan} with an output dimension $1$ and $l^{(j)}$ is a one-hot location feature.
    We note that the Tucker fusion's output dimensionality is adjustable by adding a fully-connected layer.
    \item {Step 5}: Generate a diffusion matrix $M(f_q)$ conditioned on the question $f_q$.
    \item {Step 6}: Compute the aggregated feature over cubemaps, $g_{i} = \sum_{j=1}^J\left(\sum_{v=1}^J M(f_q)_{j,v}\alpha^{(v)}\right)[l^{(j)}, g_{iq}^{(j)}]$, where we concatenate the location feature $l^{(j)}$ with the multimodal feature $g_{iq}^{(j)}$.
    \item {Step 7}: Extract a higher-level multimodal feature $g_{iq}^{(\text{higher})} =\mathcal{T}^{(\text{higher})}(g_{i}, f_q)$, where $\mathcal{T}^{(\text{higher})}$ is the Tucker fusion module with a multi-dimensional output.
    \item {Step 8}: Feed $g_{iq}^{(\text{higher})}$ in the classifier $\mathcal{C}(\cdot)$, implemented by a fully-connected layer, to predict the answer.
\end{compactitem}

\begin{figure}
\begin{center}
\includegraphics[width=.8\linewidth]{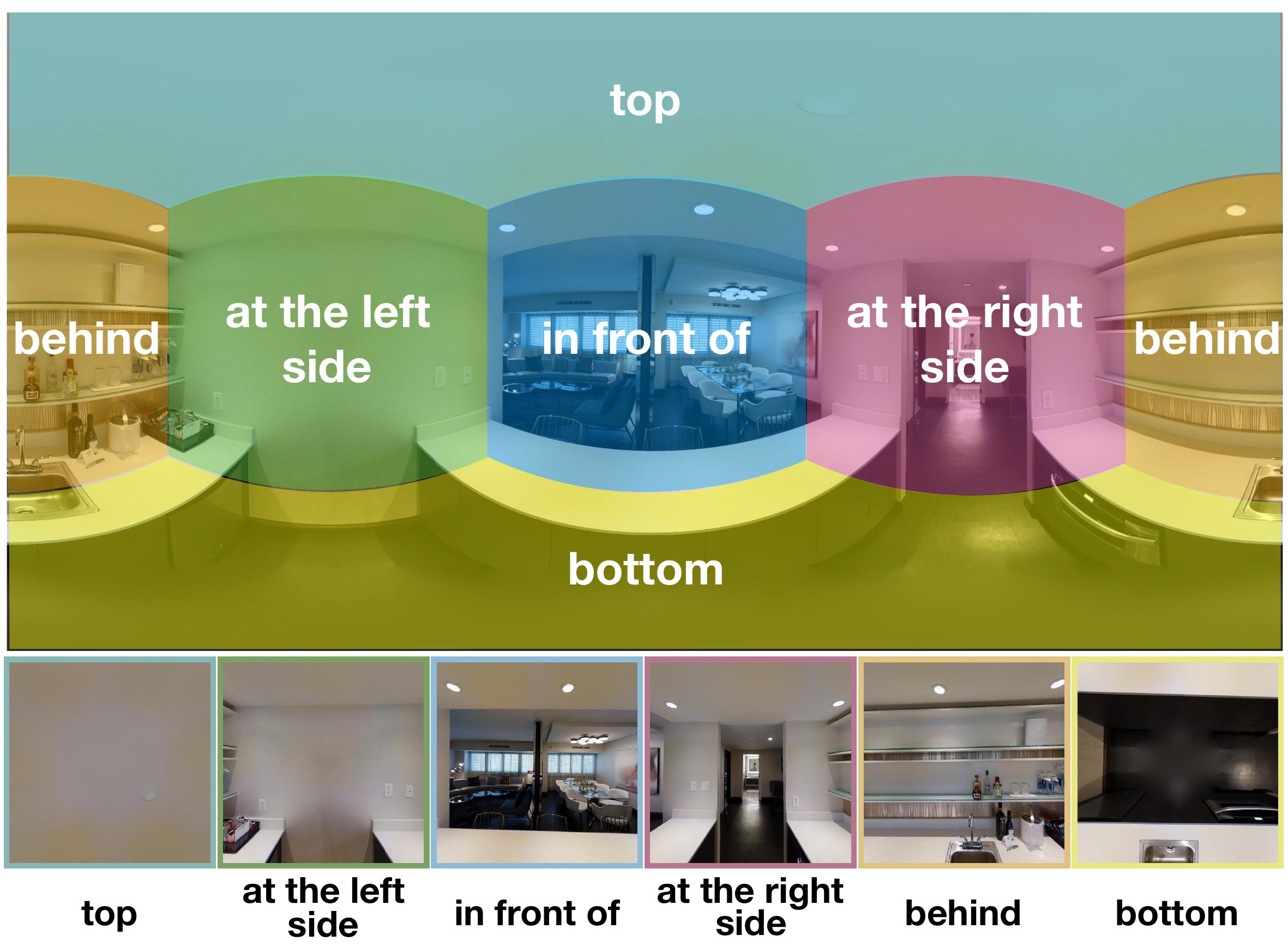}
\end{center}
\vskip -12.5pt
   \caption{\textbf{360$\mathbf{^\circ}$ image and cubemaps.}
        A 360$^\circ$ image can be represented by six cubemaps, each corresponding to a specific spatial location, to reduce the spatial distortion.}
\label{fig:tocube2}
\vspace{-2mm}
\end{figure}

\paragraph{Cubemap projection.}
The cube mapping projection is a commonly used method to project an equirectangular image onto NFOV planes~\cite{facebookcube, cheng2018cube, el2016streaming, WangACCV18}.
Specifically, there are six cube faces (top, front, left, behind, right and bottom) to fill the whole sphere as shown in Figure~\ref{fig:tocube2} in the main paper. 
We use the implementation in~\cite{cheng2018cube} to project the equirectangular images onto cubemaps.

\subsection{Additional Experimental Results}
\label{sec:addition_exp}
We provide additional comparisons on the backbone VQA models and the cubemap-based models.

\begin{table*}[!t]
    \centering
    \footnotesize
    \caption{
       \textbf{Comparison on VQA backbone models.}
        We use the MLB, MUTAN and Pythia VQA pre-trained models as the backbone in the proposed methods and evaluate the performance on our \VQATST test set.
    }
    \vskip -5pt
    \label{tab:exp_mlb}
    \begin{tabular}{c|c|c|cc|ccccc}
    \toprule
    Model       & Variants & Backbone & Overall avg & Avg by type & Scene & Exist & Counting & Property & Spatial\\ 
    \midrule
    Equirectangular-based   & \avpool      & MLB & 54.47 & 56.14 & 69.34 & 76.81 & 46.32 & 50.96 & 37.25\\
    Cubemap-based & \cubavpool & MLB & 55.03 & 56.89 & 71.41 & 76.14 & 47.72 & 52.77 & 36.42\\
    Cubemap-based & \Tucker & MLB & 57.71 & 59.07 & 69.89 & 77.23 & 46.53 & 48.24 & 53.47\\
    Cubemap-based & \TuckerM & MLB & 58.66 & 60.26 & 72.01 & 76.34 & 46.84 & 50.12 & 55.98\\
    \midrule
    Equirectangular-based   & \avpool      & MUTAN & 52.05 & 53.35 & 65.08 & 73.18 & 46.12 & 47.00 & 35.38\\
    Cubemap-based & \cubavpool & MUTAN & 53.56 & 53.56 & 69.07 & 74.86 & 46.37 & 50.72 & 35.45\\
    Cubemap-based & \Tucker & MUTAN & 54.06 & 55.29 & 65.13 & 74.59 & 45.06 & 48.20 & 43.49\\
    Cubemap-based & \TuckerM & MUTAN & 54.08 & 55.80 & 69.82 & 75.57 & 46.32 & 49.32 & 37.96\\
    \midrule
    Equirectangular-based   & -      & Pythia & 50.37 & 49.59 & 45.02 & 43.51 & 72.69 & 47.63 & 39.09 \\
    Cubemap-based & \cubavpool & Pythia & 50.90 & 51.47 & 56.37 & 43.99 & 70.92 & 48.68 & 37.37 \\
    Cubemap-based & \Tucker & Pythia & 51.88 & 51.34 & 49.84 & 46.32 & 75.48 & 47.87 & 37.21 \\
    Cubemap-based & \TuckerM & Pythia & 53.06 & 52.43 & 50.41 & 48.34 & 75.26 & 49.00 & 39.13 \\
    \bottomrule
    \end{tabular}
\end{table*}

\begin{table*}[!t]
    \centering
    \footnotesize
    \caption{
       \textbf{Comparison on the answer prediction strategies.} We compare the aggregation and fusion aggregation (cf. Figure~4 of the main paper) methods with the cubemap-based models. We report results on the \VQATST test set.
    }
    \vskip -5pt
    \label{tab:exp_fusion}
    \begin{tabular}{c|cc|cc|ccccc}
        \toprule
        Model       & Variants & Ans. Prediction & Overall avg & Avg by type & Scene & Exist & Counting & Property & Spatial\\ 
        \midrule
        Cubemap-based & \cubavpool & Aggregation & 55.03 & 56.89 & 71.41 & 76.15 & 47.72 & 52.77 & 36.42 \\
        Cubemap-based & \cubavpool & Fusion Aggregation & 54.60 & 56.23 & 69.17 & 76.22 & 46.79 & 51.72 & 37.26 \\ 
        \midrule
        Cubemap-based      & \Tucker & Aggregation & 54.12 & 54.94 & 62.97 & 75.24 & 46.22 & 46.63 & 43.62 \\
        Cubemap-based      & \Tucker & Fusion Aggregation & 57.71 & 59.07 & 69.89 & 77.23 & 46.53 & 48.24 & 53.47 \\ 
        \midrule
        Cubemap-based      & \TuckerM & Aggregation & 55.21 & 56.52 & 66.67 & 75.75 & 47.39 & 52.81 & 39.99 \\
        Cubemap-based      & \TuckerM & Fusion Aggregation & 58.66 & 60.26 & 72.01 & 76.34 & 46.84 & 50.12 & 55.98 \\
        \bottomrule
    \end{tabular}
\end{table*}

\subsection{Comparisons on Backbone VQA Models}
\label{sec:model_suppl}

We compare two VQA pre-trained models, MLB~\cite{kim2016hadamard}, MUTAN~\cite{ben2017mutan} and Pythia~\cite{singh2018pythia,singh2019TowardsVM}, as the backbone of the proposed method.
Table~\ref{tab:exp_mlb} summarizes the results of an equirectangular model, \avpool, as well as three cubemap-based models, \cubavpool, \Tucker, and \TuckerM.
We observe a similar trend as discussed in Section 5.2 of the main paper: the cubemap-based methods generally outperforms the equirectangular-based models, while the cubemap-based \TuckerM model with multi-level attention performs favorably against other variants.
Note that the state-of-the-art model on NFOV images, Pythia~\cite{singh2018pythia,singh2019TowardsVM}, does not perform better than the MLB and MUTAN models.
The possible reasons are: 1) the object detector is not generalized well to the \VQATST dataset, and 2) the cubemap project sometimes split an object into multiple parts.
These observations indicate the potential future directions on exploring the adaptive cubemap projections or object detection on $360^{\circ}$ images.

\subsection{Comparisons on Answer Prediction Strategies}\label{sec:exp_suppl}

As mentioned in Section 4 of the main paper, we fuse the multimodal feature $g_i$ with the question feature $f_q$ (which is named as Fusion Aggregation) before inputting to the classifier for predicting the answer.
Here we study another simpler strategy -- the multimodal feature $g_i$ is directly inputted into the classifier for prediction -- which is named as Aggregation.
In Table~\ref{tab:exp_fusion}, we compare these two answer prediction strategies on the cubemap-based \cubavpool, \Tucker, and \TuckerM models.

For the \cubavpool model, having a higher-level fusion degrades the performance.
However, for the \Tucker and \TuckerM models, the fusion aggregation clearly improves the overall performance.
Since both the \Tucker and \TuckerM use the location indicators as one of the features (see Step 6 in Section~\ref{sec:approach_suppl}), the fusion aggregation is necessary to associate the question to certain cubemaps so as to answer questions such as ``Which side of the tv is the pictures?'' in Figure 1 in the main paper.
We note that, as shown in Table 4 of the main paper, adding the location feature leads to notable improvements on ``spatial'' type questions.

\subsection{Additional Qualitative Results}\label{sec:result_suppl}

\begin{figure*}[ht]
\centering
  \includegraphics[width=\linewidth]{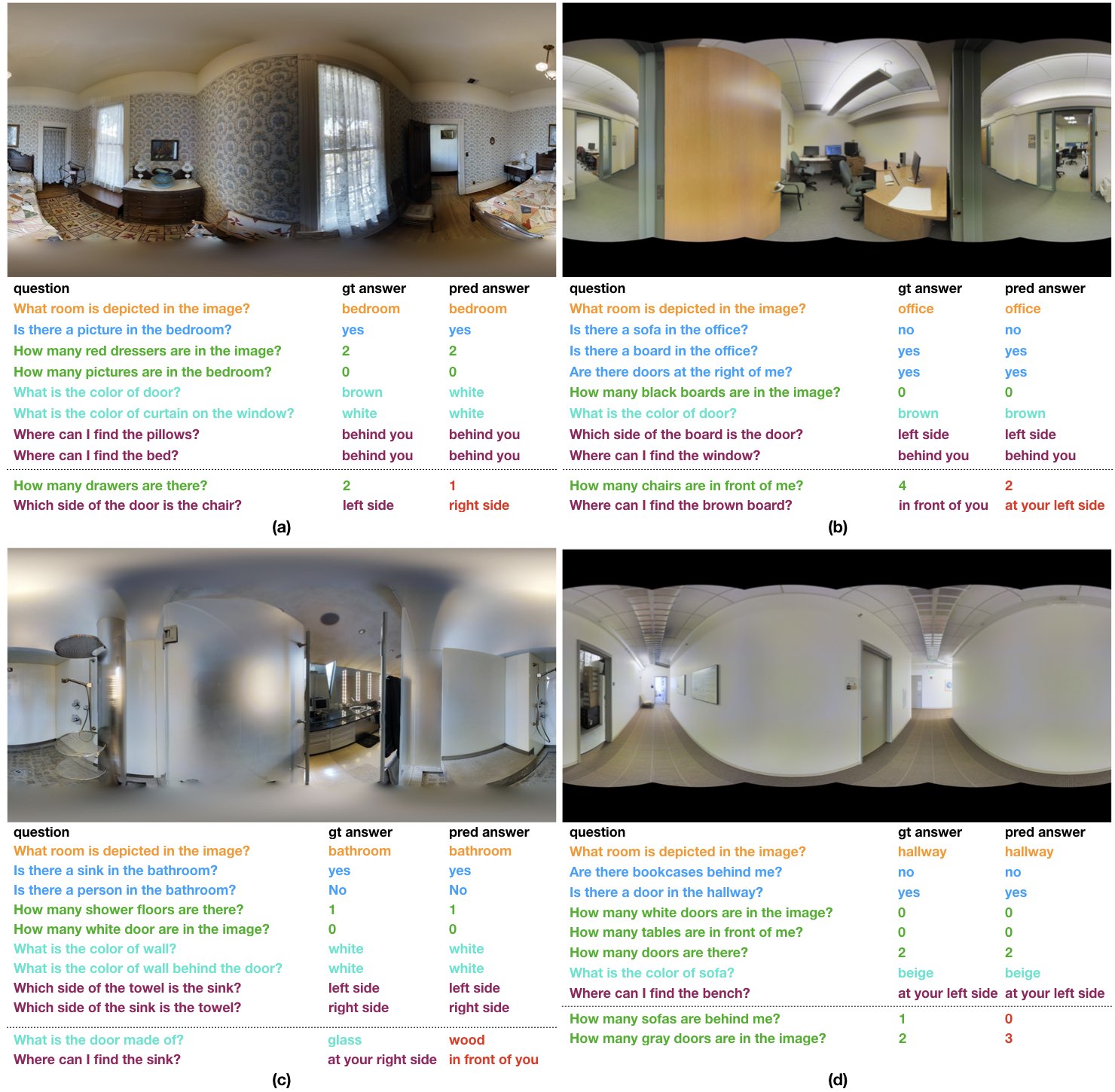}
  \caption{\textbf{Qualitative results.} We show both correct predictions and failure cases (highlighted by red font).}
  \label{fig:resultsupp}
\end{figure*}

\begin{figure*}[ht]
\centering
  \includegraphics[width=\linewidth]{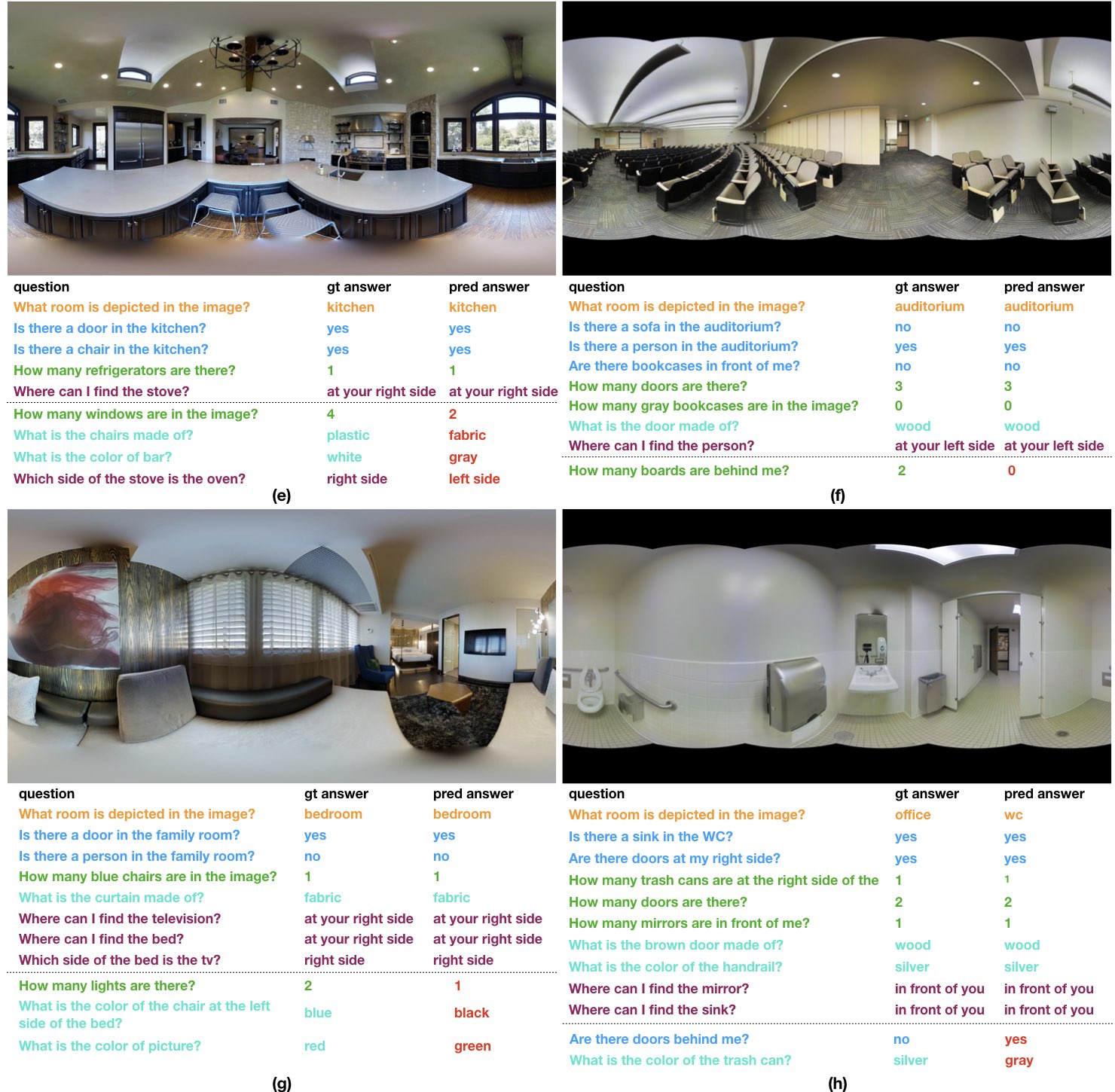}
  \caption{\textbf{Qualitative results.} We show both correct predictions and failure cases (highlighted by red font).}
  \label{fig:resultsupp2}
\end{figure*}

We provide more qualitative results using our cubemap-based model with \TuckerM in Figure~\ref{fig:resultsupp} and Figure~\ref{fig:resultsupp2}.
For each $360^\circ$ image, we show both the correct predictions and failure cases.
We observe two notable failure cases --- colors and properties --- both require accurately locating the objects according to the questions, especially for small objects.
We suggest that further improvements can be achieved by advanced object detection in 360$\mathbf{^\circ}$ images.


{\small
\bibliographystyle{ieee}
\bibliography{ref}
}

\end{document}